\documentclass[twoside,11pt]{article}

\usepackage{jmlr2e}

\usepackage[english]{babel}

\usepackage[letterpaper,top=2cm,bottom=2cm,left=3cm,right=3cm,marginparwidth=1.75cm]{geometry}

\usepackage{amsmath}
\usepackage{graphicx}
\usepackage{indentfirst}
\usepackage{bm}
 \usepackage[noabbrev,capitalize]{cleveref}
 \usepackage{natbib}
\usepackage{amsfonts}
\usepackage[dvipsnames]{xcolor}
\usepackage{amsmath}
\usepackage{amssymb}

\usepackage{algorithm2e}
\usepackage{adjustbox} 
\usepackage{tabularx}

\jmlrheading{1}{2022}{1-XX}{4/00}{10/00}{Yang Sun, Wenbin Lu and Yi-Hui Zhou*}

\ShortHeadings{Structure Maintained Representation Learning Neural Network for Causal Inference}{Sun, Lu and Zhou}
\firstpageno{1}

\begin{document}

\title{Structure Maintained Representation Learning Neural Network for Causal Inference}

\author{\name Yang Sun $^2$ \email ysun42@ncsu.edu \\
        \name Wenbin Lu $^2$  \email wlu4@ncsu.edu \\
        \name Yi-Hui Zhou $^{^*,1,2}$  \email yihui\_zhou@ncsu.edu \\
       \addr 1. Department\ of\ Biological\ Sciences\\
       2. Department\ of\ Statistics \\
       North\ Carolina\ State\ University\\
       Raleigh, NC\ 27695, USA
       }
\footnote{* corresponding author}
\editor{}

\maketitle

\begin{abstract}
Recent developments in causal inference have greatly shifted the interest from estimating the average treatment effect to the individual treatment effect. In this article, we improve the predictive accuracy of representation learning and adversarial networks in estimating individual treatment effects by introducing a structure keeper which maintains the correlation between the baseline covariates and their corresponding representations in the high dimensional space. We train a discriminator at the end of representation layers to trade off representation balance and information loss. We show that the proposed discriminator minimizes an upper bound of the treatment estimation error. We can address the tradeoff between distribution balance and information loss by considering the correlations between the learned representation space and the original covariate feature space. We conduct extensive experiments with simulated and real-world observational data to show that our proposed Structure Maintained Representation Learning (SMRL) algorithm outperforms state-of-the-art methods. We also demonstrate the algorithms on real electronic health record data from the MIMIC-III database.
\end{abstract}

\section{Introduction}

Estimating heterogenerous causal effects of a treatment has drawn increasing interests in many fields such as personalized medicine, policy making, and economics. While traditional methods focus on estimating the average causal effect on a target population, this approach is insufficient to draw inferences about differential causal effects due to the differential responses across different characteristics to a treatment. In this study, we focus on answering the question  ``which treatment works best for whom" by estimating the conditional average treatment effects (CATE) or individualized treatment effect (ITE) based on observational data.

The fundamental challenge of causal inference is that for each individual, we only observe the outcome corresponding to the assigned treatment group (factual outcome), and the other potential outcome (counterfactual outcome) under the opposite treatment option is missing (\cite{rubin2005causal,ding2017}). Therefore, the standard supervised learning approach does not apply from the prediction perspective because the counterfactual is never observed, and the actual individual causal effects remain unknown. One of the most prominent challenges to inferring the missing potential outcomes from observational data is that the treatment assignment mechanism is unknown and observational data usually suffers from selection bias, so the covariate distributions across treatment arms may be fundamentally different. For machine learning, this causes \emph{distributional shift} problem when one tries to predict, and for statistical inference, this is known as \emph{confounding}, where the confounders are variables associated with both treatment assignment and outcome, leading to biased estimation of causal effects when not properly accounted for \cite{zubizarreta2015stable}. Classical methods address covariate imbalance via propensity score methods such as matching or weighting \cite{rosenbaum1983central,kallus2020generalized,zubizarreta2015stable}. However, these methods mainly focus on estimating the average causal effect and rely on correct estimation of the propensity scores. Moreover, the popular inverse probability weighting \cite{Robins2000} may suffer from large variance when the overlap of covariate distributions is poor.  Recent developments in machine learning solve this problem via \emph{representation learning} through deep neural networks such that the  covariate distributions between treatment arms are balanced in the learned high dimensional representation space (\cite{shalit2017estimating, johansson2016learning}). However, the covariates associated with treatment assignment usually offer valuable information about final estimate of the causal effect (\cite{shi2019adapting}), and over emphasizing balance may lose such information of outcomes and harm the predictive accuracy \cite{alaa2018limits}. Therefore, representation learning faces the trade-off between achieving good balance and maintaining predictive information. Distinct from the representation learning, another popular machine learning approach directly infers ITE based on the generative adversarial nets (GANs) (\cite{goodfellow2014generative}), where the generators and discriminators are trained adversarially to learn the counterfactual outcomes and subsequently ITEs (\cite{yoon2018ganite}).  These models also showed promising results to learn complex generative distributions and operate under limited model assumptions. \textcolor{black}{In addition, the similarity preserved individual treatment effect (SITE) framework \cite{yao2018representation} learns a representation of the data that preserves local similarity and balances data distributions to minimize the influence of confounding variables. The Causal Effect Inference with Deep Latent-Variable Models (CEVAE) \cite{louizos2017causal} combines the power of variational autoencoders (VAEs) with causal graphical models to estimate individual treatment effects by learning latent representation. Deep Counterfactual Networks with Propensity-Dropout (DCN-PD) \cite{alaa2017deep} leverages dropout mechanisms within a deep neural network to estimate the propensity score and ITE with robustness and scalability. More recently,  the Treatment Effect Estimation with Disentangled Latent Factors (TEDVAE) introduces disentangled latent factors into the treatment effect estimation process, which aims to disentangle factors that affect treatment assignment from factors that influence outcomes. The works discussed above, among the important papers from the biomedical informatics venues, see  \cite{yao2019ace, ghosh2022dr, ghosh2021deep}, have significantly contributed to the progress of causal inference from observational data.}

\textcolor{black}{Recent literature for unsupervised or self-supervised representation learning discussed the importance of mutual information in acquiring meaningful representations \cite{tschannen2019mutual}, and studies have explored similarities across multiple networks by identifying neuron permutations that exhibit maximal correlation \cite{raghu2017svcca}. Inspired by these work,}
In this study, we capitalize on the success of representation learning and adversarial networks in estimating ITEs. First, we improve the predictive accuracy of representation learning by introducing a structure keeper which maintains the correlation between the baseline covariates and their corresponding representations in the high dimensional space. Second, we train a discriminator at the end of representation layers to trade off representation balance and information loss. We show that the proposed discriminator minimizes an upper bound of the treatment estimation error. We train the representation layers to fool a discriminator, which attempts to determine whether the given representations are from the treatment or the control arm. We can address the tradeoff between distribution balance and information loss by considering the correlations between the learned representation space and the original covariate feature space. We conduct experiments with simulated data and real-world observational data. The code of experiments can be found at https://github.com/SMRLNN/SMRLNN. Our proposed Structure Maintained Representation Learning (SMRL) algorithm outperforms state-of-the-art methods.

\section{Problem setup and notations}
\newcommand{\indep}{\perp \!\!\! \perp}
Consider a sample of $N$ individuals, where the treatment group ($Z = 1$) has $N_1$ subjects, and the control group ($Z=0$) contains $N_0$ subjects. We operate under the  Stable Unit Treatment Value Assumption (SUTVA) \cite{rubin1980randomization}, each subject $i$ has two potential outcomes $Y_i(1)$ and $Y_i(0)$ under treatment and control. SUTVA implies the potential outcomes of each subject are not impacted by the treatments received by others, and there is only one version of each treatment. The fundamental challenge of causal inference is that we only observe the outcome corresponding to the assigned treatment group (factual outcome), $Y_i^F = Z_i Y_i(1) + (1-Z_i)Y_i(0)$, and the other unobserved outcome (counterfactual outcome) $Y_i^{C}$ is missing. Suppose we also observe a vector of $P$ pre-treatment covariates, $ {X}_i =(X_{i1}, ..., X_{iP})^T$.  Denote the probability of receiving treatment giving covariates by $e( {X}_i) = \Pr(Z_i = 1|  {X}_i)$, i.e. \emph{propensity score}, and the conditional expectation of the potential outcome given the pre-treatment covariates with treatment $z$ by $\mu_{z}(  x)=E\{Y(z)| {X} =   x\}$ for $z=0,1$. We are interested in estimating the conditional average treatment effect (CATE) or the individual treatment effect (ITE), defined as the expected difference of potential outcomes given the pre-treatment covariates $
\tau(  x)=\mu_{1}(  x)-\mu_{0}(  x)=E\left(Y(1)-Y(0) \mid   X=  x\right)$. In addition, another causal estimand commonly of interest is the average treatment effect (ATE), $\tau_{\text{ATE}}= E_{x\sim p(x)}\{\tau(  x)\}$, where the expectation is taken on a pre-specified population of interest with covariate distribution $p(x)$.

\textcolor{black}{
Estimating the causal estimands involves the task of deducing the missing counterfactual outcomes for each individual. To ensure the identifiability of these estimands, researchers commonly rely on two well-established assumptions, as detailed by \cite{rosenbaum1983central}, (1)
Strong Ignorability (Unconfoundedness): This assumption, denoted as $Z \indep {Y(1),Y(0)}| X$, asserts that the assignment of treatment ($Z$) is independent of the potential outcomes ($Y(1)$ and $Y(0)$) given the observed covariates ($X$). Essentially, it emulates a situation akin to conditional randomization, ensuring that the treatment assignment is not influenced by hidden confounding variables; (2)
Positivity (Overlap): This assumption, expressed as $0 < e(X) < 1$, posits that for any given set of covariates ($X$), there exists a non-zero probability that an individual may belong to either the treatment or control group. In other words, it ensures that each subject has a realistic chance of receiving either treatment, preventing scenarios where certain covariate values result in an exclusive assignment to one group.
These two assumptions collectively facilitate the identification of causal estimands by addressing issues related to confounding and the distribution of treatment assignment probabilities among individuals.
}

\section{Relevant work}
Previous machine learning methods for the estimation of ITE fall into three categories. The first category directly models the outcome response surface. For example, Causal Forest (CF) (\cite{davis2017using,wager2018estimation});  Bayesian Additive Regression Trees (BART) (\cite{hill2011bayesian}); GAMLSS  (\cite{hohberg2020treatment}). The second category separately models the representation and the outcome surface such that the neural networks are encourages to learn balanced representations. For example, Treatment
Agnostic Regression Network (TARNET) and Counterfactual Regression Network (CFRNET) (\cite{johansson2016learning,shalit2017estimating, johansson2018learning}). These methods proposed two possible statistical distances to measure the  distribution discrepancies. Specifically, let  $p_{1}, p_{2}$ be two distributions over a probability space $\mathcal{S}$, the  Integral Probability Metrics (IPM) is defined as $ I P M_{\mathrm{G}}\left(p_{1}, p_{2}\right)=\sup _{g \in \mathrm{G}}\left|\int_{\mathcal{S}} g(s)\left(p_{1}(s)-p_{2}(s)\right) d s\right|$, where $g: \mathcal{S} \rightarrow \mathbb{R}$ belongs to a function family $G$.  When $G$ is the family of 1-Lipschitz functions, IPM becomes the Wasserstein distributional distances, and when $G$ is the family of norm-1 reproducing kernel Hilbert space (RKHS) functions, IPM becomes the Maximum Mean Discrepancy (MMD) distances. Penalizing IPM loss forces  the treated and control covariate distributions to be similar. The third category such as GANITE  (\cite{yoon2018ganite}) extends GAN based method by attempting to learn the counterfactual distributions and the ITE distributions.

Our work is most similar to CFRNET, but as representation learning trades off between reducing bias and maintaining predictive information, \cite{zhang2020learning}
argued that the choice of the IPMs may critically impact the model performance, and the overlap in representation space may be substantially biased. To tackle these challenges,  We propose a structure keeper that emphasizes the correlation between the learned representations and the original covariates. In addition, instead of choosing an arbitrary IPM such as Wasserstein distances or MMD, we alternately optimize a discriminator to distinguish whether the representations are transformed from the treatment or control group.

\section{Structure Maintained Representation Learning Neural Network for Causal Inference (SMRLNN)}
Adversarial representation learning capitalizes on the recent development in utilizing representation learning to achieve covariate balance in the high-dimensional space. Instead of defining specific metrics, such as Wasserstein distance or MMD distance, to measure distances between two distributions, we propose to introduce an adversarial approach where we train a discriminator to differentiate whether the learned representations $\Phi(x)$ are from the treated or control arm. Hence, the discriminator forces the representation layers to map the covariate probability space to an overlapped probability space.

\subsection{Representation Balancing}\label{ss: balancing}



Traditional balancing methods such as propensity score weighting focus on balancing the first moment condition, i.e. absolute mean difference, between two treatment groups. However, a higher moment balance is required to achieve unbiasedness when the treatment effect is heterogeneous across patients’ baseline covariates. Therefore, propensity score methods suffer from bias even when the actual propensity scores are provided. In addition, in practice, the propensity scores must be estimated from real data, and misspecification of the propensity score could result in high bias and low precision.

 In contrast, the balancing property of representation learning is guaranteed by the discriminator, which forces the distribution similarity between two treatment groups. Therefore, representation learning usually performs better under complex propensity score models and to estimate heterogeneous treatment effects.

Let $\Phi: \mathcal{X} \rightarrow \mathcal{R}^d$ be a representation function that maps from the covariate probability space $\mathcal{X}$ to a representation  space $\mathcal{R}^d$, such that the covariate distributions of different treatment arms are balanced in $\mathcal{R}^d$. The representation functions are constructed by a deep neural network, and we accomplish the goal of achieving covariate balance by adding a discriminator after the representation layers. The representation balancing discriminator $D: \mathcal{R}^d \rightarrow \mathcal{R}$ belongs to a class of classifiers  which differentiates whether the learned representations $\Phi(x)$ are from the treated or control arm. The representation layers and discriminator are trained iteratively such that the learned representations are balanced between treatment arms to be able to fool the discriminator. When we update the representation layers, the parameters of the discriminator are fixed. As a result, the penalization will make the representation layers to map samples toward the decision boundary. Therefore, traditional GANs may suffer from no loss when samples lie in a long way on the correct side of the decision boundary. To stabilize the training, we follow the the work of LSGANs \cite{mao2017least} by defining the objective of the representation balancing component as minimizing the following losses from a two-player game



\begin{align}
 L_{D}(\Phi(x|z=0),\Phi(x|z=1))=& \frac{1}{2} \mathbb{E}_{(x|z=0)} \left( D(\Phi(x)) -1\right)^2
+ \frac{1}{2} \mathbb{E}_{(x|z=1)} \left( D(\Phi(x)) +1\right)^2 \label{eq:ganloss1}\\
 L_{\Phi}(\Phi(x|z=0),\Phi(x|z=1))=& \frac{1}{2} \mathbb{E}_{(x|z=0)} \left( D(\Phi(x)) \right)^2 \label{eq:ganloss2},
\end{align}

where Equation \eqref{eq:ganloss1} is minimized with respect to the discriminator $D$, and Equation \eqref{eq:ganloss2} is minimized with respect to the representation layers. These modified losses generate more gradients by penalizing the samples lying close to the decision boundary, thus resulting in more stabilized training performance. In addition, Mao et al. \cite{mao2017least} proves that minimizing Equation \eqref{eq:ganloss1} and \eqref{eq:ganloss2} yields minimizing the Pearson $\chi^{2}$ divergence between $p(x|z=0)+p(x|z=1)$ and $2 p(x|z=1)$.



\subsection{Representation Structure Keeper}
The aim of representation layers are to balance the covariate distributions in the learned represented space, but to keep the prognostic information contained by covariates. In this section, we introduce a structure keeper on top of the representation layers based on the Representation Structure Keeper (RSK). The RSK allows for calculating correlation between two sets of variables in high dimensional space. For the given pairs of sample of covariates and their representations $\left(\left(X_{1}, \Phi(X)_{1}\right), \ldots,\left(X_{n}, \Phi(X)_{n}\right)\right)$, denote the projection of $X$ and $\Phi(X)$ in a chosen direction by
$$
P_X = W_X X, \quad P_{\Phi(X)} = W_{\Phi(X)} \Phi(X),
$$
where $W_X$ and $W_{\Phi(X)}$ are the $K\times P$ and $K\times d$ projection matrices of $X$ and $\Phi(X)$, respectively. The RSK solves the projection matrices such that the correlation defined by the top $K$ projection directions between the covariates and the representations are maximized. For example, denote the correlation matrix of $P_X$ and $P_{\Phi(X)}$ by

\begin{align}
corr(P_{X}, P_{\Phi(X)}) &= \operatorname{corr}\left(W_X X, \; W_{\Phi(X)} \Phi(X)\right) \\
&=\frac{W_X \hat{\mathbb{E}}[X \Phi(X)  ^ { \prime } ] W_{\Phi(X)}^{\prime}}
{\sqrt{W_X ( \hat{\mathbb{E}}\left[X X^{\prime}\right] +\lambda_1 I )
W_X^{\prime} W_{\Phi(X)} ( \hat{\mathbb{E}}\left[\Phi(X) \Phi(X)^{\prime}\right] +\lambda_2 I )
W_{\Phi(X)}^{\prime}}}.\label{eq:corr}
\end{align}

Then $W_X$ and $W_{\Phi(X)}$ are optimized such that

\begin{align}
L_{RSK}(x,  \Phi(X) ) &=\max _{W_X, W_{\Phi(X)}} \sum_K \text{diag}(W_X C(X, \Phi(X)) W_{\phi(X)}^{\prime}),
\label{eq:RSK}
\end{align}

with
\[
\begin{aligned}
W_X (C_{XX} +\lambda_1 I ) W_X^{\prime} &=1 \\
W_{\Phi(X)} (C_{\Phi(X)} +\lambda_2 I ) W_{\Phi(X)}^{\prime} &=1 .
\end{aligned}
\]
where $\text{diag}(C(X, \Phi(X)))$ represents the diagonal elements of the correlation matrix. Moreover, the correlation matrix $C(X, \Phi(X))$ in \eqref{eq:RSK} can be decomposed to
$$
C(X, \Phi(X))=\hat{\mathbb{E}}\left[\left(\begin{array}{l}
X \\
\Phi(X)
\end{array}\right)\left(\begin{array}{l}
X \\
\Phi(X)
\end{array}\right)^{\prime}\right]=\left[\begin{array}{ll}
C_{X X} & C_{X \Phi(X)} \\
C_{\Phi(X)X} & C_{\Phi(X)\Phi(X)}
\end{array}\right],
$$
and the corresponding Lagrangian of RSK optimization problem is

\begin{align*}
L\left(W_X, W_{\Phi(X)}\right) = \min_{\lambda_{X},\lambda_{\Phi(X)}} & \left[
-W_X C_{X \Phi(X)} W_{\Phi(X)}^{\prime}+\frac{\lambda_{X}}{2}\left(W_X (C_{X X} +\lambda_1 I ) W_X^{\prime}-1\right) \right. \\
& \left. + \frac{\lambda_{\Phi(X)}}{2}\left(W_{\Phi(X)} (C_{\Phi(X) \Phi(X)}+\lambda_2 I ) W_{\Phi(X)}^{\prime}-1\right)
\right]
\label{eq:lagrange}
\end{align*}

Therefore, our representation structure keeper is designed to optimize the objective function $L\left(\lambda, W_X, W_{\Phi(X)}\right)$, and to achieve this, one can take derivatives with respect to $x$ and $\Phi(X)$, giving
\begin{align}
&(C_{X X} +\lambda_1 I )^{-1} C_{X \Phi(X)} (C_{\Phi(X)\Phi(X)} +\lambda_2 I )^{-1} C_{\Phi(X)X} \hat{W}_X=\rho^{2} \hat{W}_X \\
&(C_{\Phi(X)\Phi(X)} +\lambda_2 I )^{-1} C_{\Phi(X)X} (C_{X X} +\lambda_1 I )^{-1} C_{X \Phi(X)} \hat{W}_{\Phi(X)}=\rho^{2} \hat{W}_{\Phi(X)},
\end{align}
where the eigenvalues $\rho^{2}$ are the squared canonical correlations and the eigenvectors $\hat{W}_X$ and $\hat{W}_{\Phi(X)}$ are the normalized canonical correlation basis vectors. Therefore $\hat{W}_X$ and $\hat{W}_{\Phi(X)}$ are the solutions of a symmetric eigenvalue problem of the form $Ax = \lambda x$.

Then our loss of representation structure keeper is:


$$
\begin{aligned}
L_{RSK}(X, \Phi(X)) &= \sum_K \text{diag} \left(\hat{W}_X \hat{\mathbb{E}}\left[X \Phi(X)  ^ { \prime } \right] \hat{W}_{\Phi(X)}^{\prime} \right).
\end{aligned}
$$




\subsection{Outcome Prediction Network}
 Let $H: \mathcal{R}^d \times\{0,1\} \rightarrow \mathcal{Y}$ be the class of outcome prediction functions defined over the representation space $\mathcal{R}^d$. We implement the standard feed-forward deep neural networks that takes the last layer of representation component and the observed treatment assignment as inputs and output an outcome prediction, $\hat{y}_{{i}}=H\left(\Phi(x_i), z_i\right)$.
 Then, the empirical mean squared error (MSE) loss function for outcome prediction is 
$L_{out}(H, \Phi)
=\frac{1}{N}\sum_{i=1}^{N}\left(\hat{y}_{{i}}-y^F_{{i}}\right)^{2}$, and the total training loss function can be expressed as
$$
L_{FL} = L_{out}(H, \Phi)+\lambda R(\Phi),
$$
where $R: \mathcal{R}^d \rightarrow \mathcal{R}$ is a regularization function and $\lambda$ is a regularization coefficient that penalizes the complex of the representation architecture.



\subsection{Algorithm}
The architecture of our proposed neural networks is summarized in Figure 1, and the optimization steps are summarized in Algorithm 1.

\begin{figure}
	\caption{ \textcolor{black}{ SMRLNN Structure: $\mathcal{X}$ represents the covariates; $z$ represents the treatment assignment; 
$\Phi: \mathcal{X} \rightarrow \mathcal{R}^d$ is a representation function;
$\hat{Y(0)}$ and $\hat{Y(1)}$ are the predicted potential outcomes;
$D$ is the Representation Balancing;
$CCA$ is the Representation Structure Keeper
 }}
	\centering
		\includegraphics[width =1\linewidth,height=4in]{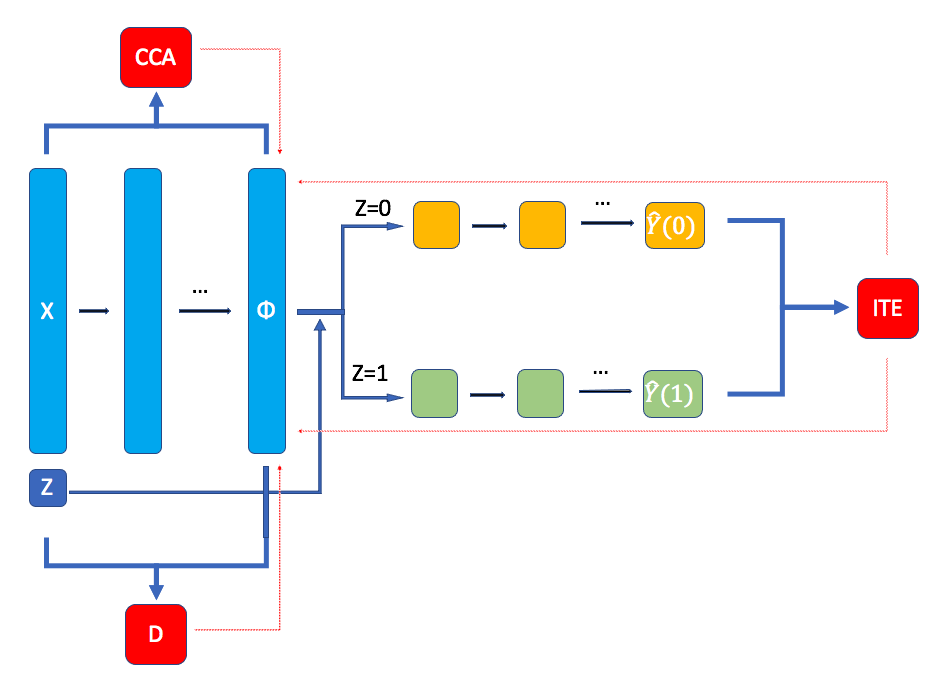}

	\label{fig:GAIN}
\end{figure}



\begin{algorithm}[hbt!]
\caption{Structure Maintained Representation Learning Neural Network for Causal Inference}\label{alg:two}

\KwData{Sample pairs $\left(x_{1}, z_{1}, y_{1}\right), \ldots,\left(x_{n}, z_{n}, y_{n}\right)$,
representation structure network $\Phi_{\textbf{W}}$ with standard normal initial weights $\textbf{W}$,
representation balancing network $D_{\textbf{U}}$ with standard normal initial weights $\textbf{U}$,
outcome network $H_{\textbf{V}}$ with standard normal initial weights $\textbf{V}$}

\KwResult{$\hat{\tau}_i = \hat{Y}_i (1) - \hat{Y}_i (0) $}

\While{not converged}{

\begin{itemize}
\item Sample mini-batch $\left\{i_{1}, i_{2}, \ldots, i_{m}\right\} \subset\{1,2, \ldots, n\}$

\item Calculate the gradients of the representation structure Keeper:
$g_{1}=\nabla_{\textbf{W}} L_{RSK}( \Phi )$

\item Calculate the gradients of the representation balancing parts:
$g_{2}=\nabla_{\textbf{U}} L(D)$,
$g_{3}=\nabla_{\textbf{W}} L( \Phi)$

\item Calculate the gradients of the outcome model:
$g_{4}=\nabla_{\textbf{V}} L_{FL}(H, \Phi)$,
$g_{5}=\nabla_{\textbf{W}} L_{FL}(H, \Phi)$

\item Update weights parameters $[\textbf{W}, \textbf{U}, \textbf{V}] \leftarrow\left[\textbf{W}-\eta\left(\alpha g_{1}+\beta g_{3}+g_{5}\right), \textbf{U}-\eta\left(g_{2}\right), \textbf{V}-\eta\left(g_{4}\right) \right]$

\item Check convergence criterion
\end{itemize}

}
\end{algorithm}










\newpage

$$\hat{\tau}_{H, \Phi}(x)=H(\Phi(x), 1)-H(\Phi(x), 0) $$
$$L_{P E H E}(H, \Phi)=\sum_{x \in \mathcal{X}}\left(\hat{\tau}_{H, \Phi}(x)-\tau(x)\right)^{2} p(x)$$

\section{Theorem}
When focusing on the Integral Probability
Metric (IPM) and Precision in Estimation of Heterogeneous Effect (PEHE), defined as $L_{P E H E}(H, \Phi)=\sum_{x \in \mathcal{X}}\left(\hat{\tau}_{H, \Phi}(x)-\tau(x)\right)^{2} p(x)$, where $\hat{\tau}_{H, \Phi}(x)=H(\Phi(x), 1)-H(\Phi(x), 0) $ is the treatment effect estimate for unit $x$, Shalit et al.\cite{shalit2017estimating} have shown that the error of PEHE is upper bounded by the sum of the expected factual loss and the IPM. We introduce $\mathcal{H}$ divergence to quantify the discriminator assessed balance condition, and show that the prediction error can be upper bounded by the sum of the expected factual loss and the $\mathcal{H}$ divergence criteria.

Let $\mathcal{D}$ denote the family of binary discriminators $D: \Phi(X) \rightarrow [0,1]$, then we define the $\mathcal{H}$ divergence \cite{ben2010theory} between two probability density distributions as:

$$
d_{\mathcal{D}}(\Phi)= \max _{ D \in \mathcal{D}}\left|\frac{1}{N_0} \sum_{x_i \in \mathcal{X}_0} D \left(\Phi(x_{i})\right)-\frac{1}{N_1} \sum_{x_j \in \mathcal{X}_1} D \left(\Phi(x_{j})\right)\right|
$$
where $\mathcal{X}_1$ and $\mathcal{X}_0$ are covariate distributions over treatment and control groups.

To facilitate the mathematical derivations, we first introduce the following definitions.
Define the expected loss for the unit and treatment pair $(x, t)$ as:
$$
\ell_{H, \Phi | z}(x)=\int_{\mathcal{Y}_z} L_Y\left(Y(z), H(\Phi(x), z)\right) p\left(Y(z) \mid x\right) d Y(z),
$$
and the maximum loss among the two treatment groups is
$
\ell_{H, \Phi}^{ max}(x) = max\left( \ell_{H, \Phi | z=0}(x), \; \ell_{H, \Phi | z=1}(x)  \right).
$



The expected factual loss and counterfactual losses of $H$ and $\Phi$ are, respectively:

\begin{align*}
    L_{F}(H,\Phi) &=\frac{1}{N} \sum_{i=1}^{N}  \ell_{H, \Phi | z=z_i} \left(x_{i}\right) p\left(x_i, z=z_i \right) \\
    L_{C}(H,\Phi) &=\frac{1}{N} \sum_{i=1}^{N}  \ell_{H, \Phi | z=z_i} \left(x_{i}\right) p\left(x_i, z=1-z_i  \right),
\end{align*}
and by the law of iterated expectations,
\begin{align*}
L_{F}(H,\Phi) &= p_0  \cdot L_{F|z=0}(H,\Phi)+p_1 \cdot L_{F|z=1}(H,\Phi)  \\
L_{C}(H,\Phi) &= p_0 \cdot L_{CF|z=1}(H,\Phi)+p_1 \cdot L_{CF|z=0}(H,\Phi) ,
\end{align*}
where $p_0 = p(z=0)$ and $p_1 = p(z=1)$, and $p\left(x,z \right) = p_0 \cdot p(x|z=0) + p_1 \cdot p(x|z=1).$

Last, the expected variance of $Y(z)$ with respect to a distribution $p(x, z)$ :



$$
\begin{aligned}
\sigma_{Y(0)}^{2}(p(x, z))&=\int_{\mathcal{X} \times \mathcal{Y}}\left(Y(0)-\mu_0(x)\right)^{2} p\left(Y(0) | x\right) p(x, z) d Y(0) d x \\
\sigma_{Y(1)}^{2}(p(x, z))&=\int_{\mathcal{X} \times \mathcal{Y}}\left(Y(1)-\mu_1(x)\right)^{2} p\left(Y(1) | x\right) p(x, z) d Y(1) d x
\end{aligned}
$$


$$
\begin{aligned}
\sigma_{Y(z)}^{2} &=\min \left\{\sigma_{Y(z)}^{2}(p(x, z)), \sigma_{Y(z)}^{2}(p(x, 1-z))\right\}, z=0,1 \\
\sigma_{Y}^{2} &=\min \left\{\sigma_{Y{(0)}}^{2}, \sigma_{Y{(1)}}^{2}\right\}
\end{aligned}
$$

\textbf{Theorem 1}
    Let \(\Phi: \mathcal{X} \rightarrow \mathcal{R}\) be a one-to-one invertible representation function, and let \(p_{\Phi}\) be the distribution induced by \(\Phi\) over \(\mathcal{R}\), i.e., \(p_{\Phi}(r | t=1)\) and \(p_{\Phi}(r | t=0)\) are the covariate distributions under treatment and control induced over \(\mathcal{R}\). Let \(L_{RSK}(X, \Phi(X))\) be the loss term associated with the Structure Keeper, which maximizes the correlation between the covariates \(X\) and their representations \(\Phi(X)\) in the learned space. We then have for any outcome prediction function \(H: \mathcal{R} \times \{0,1\} \rightarrow \mathcal{Y}\):
    \begin{eqnarray}
    L_{PEHE}(H, \Phi) \nonumber \\ & \leq &  2\left(L_{F|z=0}(H,\Phi) + L_{F|z=1}(H,\Phi)  +   d_{\mathcal{D}}(\Phi) \cdot \sum_{x \in \mathcal{X}} \ell_{H, \Phi}^{max}(x) - 2 \sigma_{Y}^{2}\right) \\ 
    &-& \lambda \cdot L_{RSK}(X, \Phi(X)) \nonumber,
    \end{eqnarray}
    where \(\lambda > 0\) is a regularization parameter that controls the influence of the Structure Keeper on the overall loss.

See proof in appendix.

\textbf{Remark.} Theorem 1 establishes the lower bound of PEHE for any outcome prediction function using representation learning, when the distance of representation space and the original covariate space is measured by the $\mathcal{H}$ divergence. The first two terms in (9) relate to the outcome prediction error, and are optimized by the typical supervised learning process using neural networks. The third term involves the product of the  treatment distribution distance  quantified by  $\mathcal{H}$ divergence and the maximum expected loss among the two treatment groups. While the  maximum expected loss is fixed given the optimal outcome prediction function $H$ and the representation function $\Phi$, our proposed algorithm minimizes the $\mathcal{H}$ divergence via optimization of the discriminator introduced in Section \ref{ss: balancing}. Theorem 1 lays the theoretical foundation to ensure  the proposed algorithm to provide low prediction error of the ITE measured by PEHE, and we further validate the performance via synthetic and real data simulations in Section \ref{s:simu} and \ref{s:real}.

\section{Simulation Study}
\label{s:simu}
We design  simulations studies to compare a number of state-of-art machine learning methods that are popular for estimating the potential outcomes. The methods under comparison are Causal Forest (CF), Bayesian Additive Regression Trees (BART), Treatment
Agnostic Regression Network (TARNET) and Counterfactual Regression Network (CFRNET), and Generative Adversarial Nets for inference of Individualized Treatment Effects (GANITE). CF is a nonparametric random forests based algorithm that provides desirable asymptotic properties, and serves as a popular benchmark method   (\cite{davis2017using}); BART applies a Bayesian modeling approach by building a sum-of-trees model (\cite{hill2011bayesian}); TARNET applies representation learning without penalizing the representation balance; CFRNET incorporates the IPM loss into representation leaning (\cite{johansson2018learning}); and GANITE is a GAN based method to learn the counterfactual distributions (\cite{yoon2018ganite}).

\subsection{Data Generation Process}
 We consider various combinations of sample sizes and outcome surfaces to examine the performance of the afore mentioned methods. In total, there are $4 \text{ (sample size)}\times 3\text{ (outcome model)}=12$ simulation scenarios.

We generate $N\in \{200, 300, 500, 1000\}$ patients, with $P = 15$ covariates that are multivariate normal distributed as $
X_{i}=(X_{i1},\dots, X_{iP}) \sim \mathcal{N}\left(0, \sigma^{2}\left[(1-\rho) I_{P}+\rho 1_{P} 1_{P}^{T}\right]\right)
$,
where $\sigma^2=1$ is the marginal variance  and $\rho=0.3$ controls the correlation between the covariates for $i = 1,\dots, N$. For each subject, the observed treatment assignment $Z_i$ is simulated from a Bernoulli distribution $Z_i \sim Bernoulli(e( {X}_i))$, where $e( {X}_i)$  is the propensity score. We assume the baseline covariates serve as confounders and the propensity score model is
$$
\text{logit}[e(  X_i)] =  X_i^T   \alpha, \quad    \alpha \sim \operatorname{Unif}\left([-1, 1]^{P} \right).
$$
The realized values of the regression coefficients are  $\alpha = $ (0.8, -0.8, -1, -0.8,  0.2, -0.4,  1,  0.6,  0.2,  0.6, -0.2, -0.4, -1,  0.6,  0.4), resulting in approximately $50\%$ of the subjects being in the treatment group.

\begin{figure}
		\caption{\textcolor{black}{ Propensity Score distribution by treatment group: red represents the treated group; blue represents the control group
	}}
	\centering
		\includegraphics[width =0.8\linewidth,height=3in]{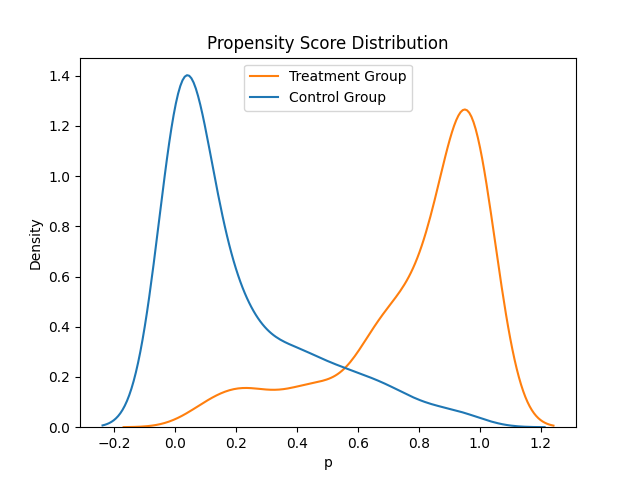}

	\label{fig:GAIN}
\end{figure}

 Overall, the observed outcome can be expressed as
\begin{align}
Y_i &= \mu_0 (  X_i) + Z_i \cdot c(  X_i) + \epsilon_i , \quad  \epsilon_i \overset{iid}{\sim} \mathbb{N} (0,1),
\end{align}
where $\mu_0 (  X_i)$ is the conditional expectation of the potential outcome under control,  $c(  X_i)$ is the  individual treatment effect that we are interested to estimate, and $\epsilon_i$ represents the random noise. This model implies that the conditional expectation of the potential outcome under treatment is $\mu_1 (  X_i)=  \mu_0 (  X_i) + c(X_i)$. To assess the robustness of different methods, we consider three outcome generation processes that satisfy linear, piece-wise linear and non-linear surfaces, separately.

In \emph{outcome model 1},  we assume a complex linear relationship motivated by Susan Athey at el. \cite{athey2017estimating}. Specifically,
$$
\begin{aligned}
\mu_{0}(  X_i) &=  X_i^{T}   \beta_{0}, \text { with }   \beta_{0} \sim \operatorname{Unif}\left([1,2]^{P}\right) , \\
c(  X_i) & \sim X_i^{T}   \beta_{1} + 2, 
\end{aligned}
$$
The realized values of the outcome regression coefficients are $\beta_0 =$ (1.2, 1.1, 1.0, 1.8, 1.6, 2.0, 1.2, 1.3, 1.4, 1.1, 1.5, 1.1, 1.1, 1.0, 1.7), $\beta_1 =$ (1.5, 1.0, 1.9, 2.0, 1.5, 2.0, 2.0, 1.7, 2.0, 1.5, 1.4, 1.6, 1.9, 1.2, 1.2).


In \emph{outcome model 2},  we assume a piece-wise linear relationship motivated by Kunzel et al. \cite{kunzel2019metalearners}:
$$
\begin{aligned}
\mu_{0}(  X_i) &=  X_i^{T} \beta_{0}, \text { with } \beta_{0} \sim \operatorname{Unif}\left([-5,5]^{P}\right) , \\
c(  X_i) &=  0.5\mathbb{I}\left(X_{i1}>0.5\right)+ \mathbb{I}\left(X_{i2}>0.3\right) +2\mathbb{I}\left(X_{i3}>0, X_{i4}>0.2\right)
\end{aligned}
$$
where $\mathbb{I}(\cdot)$ stands for the indicator function, and the realized values of the outcome regression coefficients are $\beta_0 =$ (-5,  4,  3, -2, -2, -5, -2,  2, -2,  1, -3, -5,  4,  5, -4).

In \emph{outcome model 3},  we assume a complex non-linear relationship motivated by Kang and Schafer  \cite{kang2007demystifying}:


$$
\begin{aligned}
\mu_{0}(  X_i) &=  X_i^{T}   \beta_{0},  \\
\mu_{1}(  X_i) &= \exp \left( (X_{i} + W) \beta_0   \right) \\
c(  X_i) & = \mu_{1}(  X_i) - \mu_{0}(  X_i)
\end{aligned}
$$

where $W$ is an offset matrix of the same dimension as $X_i$ with every value equal to 0.5, $\beta_{0}$ is a vector of regression coefficients (0, 0.1, 0.2, 0.3, 0.4) randomly sampled with probabilities (0.6, 0.1, 0.1, 0.1, 0.1). The realized values of the outcome regression coefficients are $\beta_0 =$ (0.1, 0.2, 0.3, 0.1, 0, 0.3, 0, 0, 0, 0, 0, 0.1, 0, 0, 0.3).

Under each setting, we simulate 100 repeated data sets and evaluate the performance of different methods by the expected precision in estimation of heterogeneous effect (PEHE) \cite{hill2011bayesian},
$$
\epsilon_{PEHE}=\frac{1}{n} \sum_{i=1}^{n}\left( \widehat \mu_1(  X_i)-\widehat\mu_0(  X_i)-\left(  \mu_1(  X_i)-\mu_0(  X_i)\right) \right)^{2},
$$
where $ \widehat \mu_0(  X_i), \widehat\mu_1(  X_i)$ are the estimated means from model, and $  \mu_0(  X_i),\mu_1(  X_i)$ are the underlying true conditional means under control and treatment group. In addition of the estimation of individual treatment effect, we also evaluate the empirical absolute bias of ATE on the overall sample,
$$
\epsilon_{ATE}= | \frac{1}{n} \sum_{i=1}^{n}\widehat \mu_1(  X_i)-\widehat\mu_0(  X_i)-  ATE  |,
$$
where the true ATE is obtained from calculating the average treatment effect of a super population with $100,000$ simulated subjects.

\subsection{Simulation Results}
\textcolor{black}{Table \ref{t:ablation} presents the performance of ITE and ATE estimation using various versions of the SMRLNN model. These versions are as follows: \\
SMRLNN-v0: SMRLNN without both Structure Keeper and Representation Balancing;\\
SMRLNN-v1: SMRLNN without Structure Keeper;\\
SMRLNN-v2: SMRLNN without Representation Balancing.\\
Across all sample sizes, SMRLNN consistently yields the smallest PEHE and the smallest error in estimating ATE. Following SMRLNN in terms of performance are SMRLNN-v2, SMRLNN-v1, and SMRLNN-v0. These results indicate that minimizing the distance in covariate distribution through the discriminator has the most significant impact in reducing estimation error, while preserving information through the structure keeper plays a comparatively lesser role.}

\begin{table}[]
\caption{Performance comparison between SMRLNN and its ablation methods as the sample sizes are varied with respect to $\epsilon_{PEHE}$ and $\epsilon_{ATE}$. The Monte Carlo SD is shown after ± . \\
}
\begin{tabular}{llllll} \hline

Metric & N    & SMRLNN-v0               & SMRLNN-v1      & SMRLNN-v2      & SMRLNN        \\
\hline
                 & 200  & 2.02 ± 0.23 & 1.85 ± 0.23 & 1.58 ± 0.21 & \textbf{1.47 ± 0.18} \\
                & 300  & 1.67 ± 0.24 & 1.60 ± 0.22 & 1.48 ± 0.28 & \textbf{1.46 ± 0.17} \\
$\epsilon_{PEHE}$                   & 500  & 1.49 ± 0.15 & 1.41 ± 0.16 & 1.36 ± 0.17 & \textbf{1.33 ± 0.18} \\
                & 800  & 1.14 ± 0.09 & 1.09 ± 0.09 & 1.04 ± 0.10 & \textbf{1.02 ± 0.11} \\
                 & 1000 & 1.10 ± 0.11 & 1.08 ± 0.11 & 1.04 ± 0.10 & \textbf{1.01 ± 0.11} \\
\multicolumn{1}{l}{} &      &             &             &             & \textbf{}            \\
                 & 200  & 0.27 ± 0.11 & 0.26 ± 0.10 & 0.20 ± 0.07 & \textbf{0.18 ± 0.06} \\
              & 300  & 0.27 ± 0.11 & 0.26 ± 0.11 & 0.24 ± 0.11 & \textbf{0.19 ± 0.08} \\
$\epsilon_{ATE}$                    & 500  & 0.22 ± 0.09 & 0.20 ± 0.07 & 0.19 ± 0.07 & \textbf{0.12 ± 0.04} \\
                 & 800  & 0.19 ± 0.08 & 0.15 ± 0.05 & 0.17 ± 0.06 & \textbf{0.13 ± 0.06} \\
                 & 1000 & 0.16 ± 0.06 & 0.14 ± 0.04 & 0.12 ± 0.04 & \textbf{0.12 ± 0.03} \\
\hline
\end{tabular}
\label{t:ablation}
\end{table}



\begin{table}[]
\caption{Performance comparison between SMRLNN and state-of-the-art methods as the outcome model and sample sizes are varied with respect to $\epsilon_{PEHE}$. The Monte Carlo SD is shown after ± .}
\setlength{\tabcolsep}{6pt}
  \begin{adjustbox}{angle=90,center,width=6in} 

\begin{tabular}{lllllllll}
\hline
Model & N    & SMRLNN               & TARNET      & CFRNET      & CF          & BART                 & GANITE      & CEVAE       \\
1 & 200  & \textbf{1.43 ± 0.07} & 2.77 ± 0.09 & 2.17 ± 0.06 & 7.71 ± 0.77 & 5.96 ± 0.63          & 9.82 ± 0.76 & 3.34 ± 0.23 \\
1 & 300  & \textbf{1.32 ± 0.11} & 1.70 ± 0.05 & 1.46 ± 0.05 & 7.23 ± 0.70 & 4.88 ± 0.45          & 9.18 ± 0.57 & 2.75 ± 0.16 \\
1 & 500  & \textbf{0.93 ± 0.07} & 1.04 ± 0.02 & 0.99 ± 0.02 & 6.81 ± 0.52 & 3.64 ± 0.26          & 7.31 ± 0.51 & 2.26 ± 0.14 \\
1 & 1000 & \textbf{0.70 ± 0.04} & 0.82 ± 0.02 & 0.75 ± 0.02 & 6.15 ± 0.39 & 2.57 ± 0.15          & 6.46 ± 0.36 & 1.32 ± 0.11 \\
  &      &                      &             &             &             &                      &             &             \\
2 & 200  & \textbf{1.56 ± 0.04} & 2.22 ± 0.04 & 1.99 ± 0.03 & 2.70 ± 0.84 & 1.87 ± 0.25          & 2.57 ± 0.18 & 2.47 ± 0.13 \\
2 & 300  & \textbf{1.51 ± 0.02} & 1.68 ± 0.02 & 1.63 ± 0.02 & 2.43 ± 0.56 & 1.68 ± 0.22          & 2.55 ± 0.16 & 1.97 ± 0.09 \\
2 & 500  & \textbf{1.35 ± 0.03} & 1.43 ± 0.01 & 1.42 ± 0.01 & 2.22 ± 0.41 & 1.37 ± 0.17          & 2.43 ± 0.18 & 1.69 ± 0.06 \\
2 & 1000 & 1.15 ± 0.02          & 1.29 ± 0.01 & 1.27 ± 0.01 & 1.97 ± 0.28 & \textbf{1.02 ± 0.13} & 2.09 ± 0.16 & 1.54 ± 0.04 \\
  &      &                      &             &             &             &                      &             &             \\
3 & 200  & \textbf{1.16 ± 0.13} & 2.36 ± 0.07 & 2.08 ± 0.07 & 2.05 ± 0.67 & 2.06 ± 0.66          & 2.87 ± 0.72 & 2.58 ± 0.27 \\
3 & 300  & \textbf{1.31 ± 0.12} & 1.95 ± 0.06 & 1.66 ± 0.05 & 1.97 ± 0.58 & 1.91 ± 0.58          & 2.61 ± 0.60 & 2.09 ± 0.23 \\
3 & 500  & \textbf{1.25 ± 0.22} & 1.49 ± 0.06 & 1.28 ± 0.06 & 1.88 ± 0.47 & 1.77 ± 0.44          & 2.27 ± 0.47 & 1.65 ± 0.21 \\
3 & 1000 & \textbf{1.01 ± 0.10} & 1.17 ± 0.07 & 1.04 ± 0.06 & 1.83 ± 0.45 & 1.61 ± 0.42          & 2.08 ± 0.45 & 1.26 ± 0.22 \\\hline
\end{tabular}
  \end{adjustbox}
  \label{t:sim_pehe}

\end{table}



\begin{table}[]
\caption{Performance comparison between SMRLNN and state-of-the-art methods as the outcome model and sample sizes are varied with respect to $\epsilon_{ATE}$. The Monte Carlo SD is shown after ± . The underlying true ATE of the three models are 2, 1.766, 3.306 respectively.}
\setlength{\tabcolsep}{4pt}
  \begin{adjustbox}{angle=90,center,width=6in} 

\begin{tabular}{llllllllll}
\hline
M & N    & SMRLNN               & TARNET      & CFRNET      & CF                   & BART        & GANITE      & DR          & CEVAE       \\\hline
M & N    & SMRLNN               & TARNET      & CFRNET      & CF                   & BART        & GANITE      & DR                   & CEVAE       \\
1 & 200  & 0.20 ± 0.07          & 0.53 ± 0.06 & 0.40 ± 0.05 & 0.56 ± 0.98          & 0.21 ± 1.04 & 0.67 ± 0.26 & \textbf{0.14 ± 1.09} & 0.64 ± 0.13 \\
1 & 300  & 0.18 ± 0.06          & 0.28 ± 0.03 & 0.29 ± 0.03 & 0.26 ± 0.80          & 0.08 ± 0.80 & 0.55 ± 0.21 & \textbf{0.01 ± 0.91} & 0.35 ± 0.08 \\
1 & 500  & \textbf{0.09 ± 0.04} & 0.25 ± 0.03 & 0.26 ± 0.03 & 0.20 ± 0.68          & 0.05 ± 0.62 & 0.43 ± 0.20 & \textbf{0.02 ± 0.72} & 0.31 ± 0.11 \\
1 & 1000 & \textbf{0.06 ± 0.02} & 0.20 ± 0.03 & 0.22 ± 0.03 & 0.21 ± 0.55          & 0.07 ± 0.46 & 0.39 ± 0.20 & 0.07 ± 0.45          & 0.26 ± 0.10 \\
  &      &                      &             &             &                      &             &             &                      &             \\
2 & 200  & \textbf{0.29 ± 0.12} & 0.45 ± 0.06 & 0.38 ± 0.05 & 1.59 ± 1.33          & 0.30 ± 0.62 & 1.69 ± 0.62 & \textbf{0.23 ± 0.66} & 0.43 ± 0.18 \\
2 & 300  & \textbf{0.18 ± 0.07} & 0.32 ± 0.04 & 0.31 ± 0.04 & 1.50 ± 0.88          & 0.27 ± 0.40 & 1.63 ± 0.64 & 0.19 ± 0.44          & 0.37 ± 0.13 \\
2 & 500  & \textbf{0.15 ± 0.05} & 0.24 ± 0.03 & 0.25 ± 0.03 & 1.30 ± 0.63          & 0.18 ± 0.29 & 1.33 ± 0.55 & \textbf{0.13 ± 0.25} & 0.29 ± 0.14 \\
2 & 1000 & \textbf{0.08 ± 0.02} & 0.20 ± 0.03 & 0.20 ± 0.03 & 1.10 ± 0.46          & 0.13 ± 0.16 & 1.24 ± 0.51 & 0.12 ± 0.14          & 0.27 ± 0.09 \\
  &      &                      &             &             & \textbf{}            &             &             &                      &             \\
3 & 200  & 0.14 ± 0.06          & 0.39 ± 0.04 & 0.33 ± 0.04 & \textbf{0.04 ± 0.28} & 0.20 ± 0.28 & 0.45 ± 0.27 & 0.06 ± 0.31          & 0.44 ± 0.14 \\
3 & 300  & 0.13 ± 0.05          & 0.27 ± 0.03 & 0.25 ± 0.03 & \textbf{0.05 ± 0.23} & 0.15 ± 0.23 & 0.43 ± 0.29 & 0.13 ± 0.31          & 0.34 ± 0.11 \\
3 & 500  & 0.11 ± 0.05          & 0.22 ± 0.03 & 0.21 ± 0.03 & \textbf{0.04 ± 0.20} & 0.08 ± 0.17 & 0.28 ± 0.22 & 0.06 ± 0.19          & 0.27 ± 0.13 \\
3 & 1000 & 0.07 ± 0.03          & 0.17 ± 0.02 & 0.18 ± 0.02 & \textbf{0.03 ± 0.12} & 0.05 ± 0.12 & 0.24 ± 0.19 & 0.05 ± 0.14          & 0.22 ± 0.11   \\\hline
\end{tabular}
  \end{adjustbox}
    \label{t:sim_ate}

\end{table}

\begin{table}[]
\caption{Performance comparison between SMRLNN and state-of-the-art methods as the numbers of covariates is increased with respect to $\epsilon_{PEHE}$. The Monte Carlo SD is shown after ± .}
\begin{tabular}{lllllrl}
\hline
P   & N   & SMRLNN               & TARNET      & CFRNET      & \multicolumn{1}{l}{CF} & BART        \\ \hline
50  & 200 & \textbf{1.56 ± 0.13} & 2.19 ± 0.12 & 2.68 ± 0.15 & 2.12 ± 0.20            & 1.66 ± 0.17 \\
100 & 200 & \textbf{1.75 ± 0.14} & 2.33 ± 0.14 & 2.90 ± 0.16 & 2.23 ± 0.21            & 2.55 ± 0.20 \\
200 & 200 & \textbf{2.45 ± 0.16} & 3.31 ± 0.35 & 3.49 ± 0.18 & 2.95 ± 0.37            & 3.69 ± 0.38 \\
400 & 200 & \textbf{3.44 ± 0.22} & 7.37 ± 0.54 & 7.17 ± 0.61 & 4.59 ± 0.44            & 5.85 ± 0.40 \\
800 & 200 & \textbf{4.58 ± 0.31} & 8.20 ± 0.44 & 5.37 ± 0.36 & 4.90 ± 0.60            & 6.89 ± 0.54 \\
\hline
\end{tabular}
\label{t:hign_dim}
\end{table}

Table \ref{t:sim_pehe} shows the performance of ITE estimation from different methods evaluated by PEHE. As the sample size increases from 200 to 1000, the PEHE of all methods monotonically decreases in all three outcome models.  Overall, SMRLNN results in the smallest PEHE and Monte Carlo standard deviation across all the methods under comparison, substantially outperforming CF and GANITE. The difference is most pronounced when the outcome model is linear. Specifically, while the PEHE of GANITE ranges from $6.46$ to $9.82$, the PEHE of SMRLNN is only $0.70$ to $1.43$. Under piese-wise liner and nonlinear outcome surfaces, the PEHE of SMRLNN is about half of GANITE. The PEHE of CFRNET is slightly better than TARNET and CEVAE, showing penalizing the representation imbalance improves  model performance when baseline covariates are imbalanced.  Under Outcome model 1, CFRNET and TARNET outperform BART, while their performances are comparable under Outcome model 2 and 3. Last but not least, the  Monte Carlo SD of CF, BART and GANITE are significantly larger than the representation learning based methods such as SMRLNN, TARNET , CFRNET and CEVAE.

Table \ref{t:sim_ate} shows the performance of ATE estimation corresponding to Table \ref{t:sim_pehe}. Similar to the trends observed in Table \ref{t:sim_pehe}, the absolute bias of ATE estimation decreases as sample size increases. Our proposed method SMRLNN achieves the smallest bias and Monte Carlo standard deviation in comparison with other methods. The improvement of SMRLNN on the ATE estimation is not as significant as the improvement of the ITE since our method is not designed for ATE estimation. While CF and GANITE remains having the largest bias, BART achieves comparable ATE bias with SMRLNN. The bias of TARNET, CFRNET and CEVAE lies between SMRLNN and CF under Outcome model 1 and 2, but CF results in the smallest ATE bias when outcome is nonlinear. Again, the  variability of CF, BART and GANITE are significantly larger than the representation learning based methods.

Table \ref{t:hign_dim} shows the performance of ITE estimation in high-dimension from different methods evaluated by PEHE. The sample size is fixed as 200, as the number of covariates increase from 50 to 800, the PEHE of all methods monotonically decreases in all three outcome models. Overall, SMRLNN results in the smallest PEHE and Monte Carlo standard deviation across all the methods under comparison.

The simulation results demonstrate that SMRLNN robustly outperforms state-of-the-art methods in terms of individual treatment effect estimations under a range of the examined scenarios. It also shows superiority in the estimation of ATE under linear and piece-wise linear outcome surfaces, as well as maintains small variance.

\section{Real Data Experiments}
\label{s:real}
In this section, we demonstrate the performance of SMRLNN architecture on real data experiments. The performance of causal inference methods is usually evaluated by a hybrid of real data variables and synthesized outcomes.
\subsection{Infant Health and Development Program Dataset}
The Infant Health and Development Program (IHDP) dataset \cite{hill2011bayesian} is a popular benchmark for evaluation of causal inference methods. IHDP is a randomized trial aiming to evaluate the efficacy of high-quality child care on premature infants. An observational study was recreated from IHDP by removing a non-random portion from the subjects, resulting in 139 children in the treatment group, and 608 in the control. Following \cite{shalit2017estimating}, we use 25 pre-treatment covariates and the simulated response surface B of  \cite{hill2011bayesian} For the IHDP data, since the outcome surface is known and both factual and counterfactual outcomes are simulated, we are able to compute the true ITE, and then evaluate using the empirical PEHE and ATE. We report the in-sample and out-of-sample performance on 100 replications of the data.
\subsection{Jobs Dataset}
Next, we evaluate various methods on another widely used benchmark based on a real-world dataset, Jobs \cite{lalonde1986evaluating, shalit2017estimating}, which combines a randomized trial with observational data such that training can be conducted on both, but only the randomized data is used for evaluation. The Jobs data includes a binary outcome, 8 covariates, with the randomized trial having 297 treated and 425 controls, and the observational data having 2490 controls. For the Jobs data, since only factual outcomes are available but the testing set comes from a randomized controlled trial (RCT), empirical policy risk is used to evaluate the average loss on the randomized subset of Jobs. Policy Risk $\left(\mathcal{R}_{p o l}(\pi)\right)$ can be defined as the average loss in value when treating according to the policy implied by an ITE estimator:
$$
R_{p o l}(\pi)=\frac{1}{N} \sum_{i=1}^{N}\left[1-\left(
\frac{1}{\left|\Pi_{1}  \cap T \right|} \sum_{i \in \Pi_{1}  \cap T} \hat{y}_i(1) \times \frac{\left|\Pi_{1} \cap E\right|}{|E|}
+
\frac{1}{\left|\Pi_{0} \cap C \cap E\right|} \sum_{i \in \Pi_{0} \cap C \cap E} \hat{y}_i(0) \times \frac{\left|\Pi_{0} \cap E\right|}{|E|}
\right)\right],
$$
where $\hat{y}_i(z)$ is the predicted  probability of employment under treatment $z$, $\Pi_{z}=\{i: z=\arg \max \hat{\textbf{y}}_i(z)\}$ is the set of randomized subjects whose predicted potential probability is larger under treatment $z$, $E$ represents the set of subjects in the RCT,  $C$ is the set of control subjects, $T$ is the set of treated, and $|\cdot|$ represents the sample size of a set. In addition,  we evaluate the empirical absolute bias of ATT on the randomized set $E$:
$$
\epsilon_{ATT}=\left| \frac{1}{|T|} \left( \sum_{i\in T }\widehat \mu_1(  X_i)-\widehat\mu_0(  X_i)\right)-ATT \right|,
$$
where $ATT= |T|^{-1}\sum_{i\in T}y_i- |C\cap E|^{-1}\sum_{i\in C\cap E}y_i$ is the average treatment effect for the treated calculated from the RCT set.

\subsection{MIMIC-III Sepsis Cohort Dataset}

The Medical Information Mart for Intensive Care-III (MIMIC-III) \cite{johnson2016mimic} is a public critical care database which includes all patients admitted to the ICUs of Beth Israel Deaconess Medical Center in Boston, MA from 2008 - 2012. The database contains information about patients' demographics, diagnosis codes, laboratory tests, vital signs, and clinical events, for over 350 million values across various sources of data (\cite{sun2022machine}).
We evaluate the treatment effect of mechanical ventilation on in-hospital mortality in adult patients fulfilling the international consensus sepsis-3 criteria. Of the 20,225 eligible admissions, 4,210 (20.8\%) received mechanical ventilation, and 1,208 (28.7\%) experienced in-hospital deaths. We pre-specify 47 baseline covariates based on clinical knowledge, including demographics, Elixhauser premorbid status, vital signs, laboratory values, fluids and vasopressors received and fluid balance (\cite{komorowski2018artificial}). Data variables with multiple measurements are recorded at the time of sepsis diagnosis. Table \ref{t:baseline} presents the baseline characteristics of these covariates. Significant imbalance is observed in many covariates, with mechanical ventilation patients being on average having more severer symptoms as evidenced by larger initial SOFA score, elixhauser score, SGOT, SGPT, IV fluid intake, and Urine output over 4 hours.

We pre-specify 47 baseline covariates based on clinical knowledge, including demographics, Elixhauser premorbid status, vital signs, laboratory values, fluids and vasopressors received and fluid balance (\cite{komorowski2018artificial}).
Table \ref{t:baseline} presents the baseline characteristics of these covariates.

 Propensity score matching (PSM) is a broadly used method for causal inference on real data.
In the literature, the individual treatment effect $\tau(X_i)$ is usually approximated by a matched pair approach, i.e., find a nearest neighbor of unit i and take the difference in outcomes of the pair as the approximated “true” ITE as described in \cite{shalit2017estimating}. In this paper, we fitted a logistic regression propensity score mode with the 25 covariates to estimate the propensity score. For each patient receiving mechanical ventilation (MV), we find a matched pair using the k-nearst neighbor method without replacement, and then take the difference in outcomes of the pair. We evaluate different methods using PEHE based on this approximated ground truth ITE.

\subsection{Twins}
\textcolor{black}{The Twins dataset is meticulously curated and originates from the ``Linked Birth/Infant Death Cohort Data" provided by the National Bureau of Economic Research (NBER). Only twin pairs that share the same gender and have a birth weight below 2000 grams are included from year 1989 to 1991. This deliberate selection ensures a focus on a specific subset of twin births, which can be especially valuable for research aimed at understanding various aspects of birth outcomes, health disparities, and treatment effects. Inspired by \cite{louizos2017causal}, we use treatment labels (`t=0' for the lighter twin and `t=1' for the heavier twin) and utilize the mortality rate of each twin during their first year of life as a pivotal metric for evaluating treatment outcomes. To simulate the presence of selection bias, we intentionally opt to observe only one of the twins in each pair concerning the covariates associated with each unit, as follows:
$t_i \mid x_i \sim \operatorname{Bernoulli}\left(\sigma\left(w_0^T x+w_h\right)\right)$, where $w_0 \sim \mathcal{N}(0,0.1 \cdot I)$ and $w_h \sim \mathcal{N}(2,0.1)$}


\subsection{Results on the real data }
\textcolor{black}{
The evaluation of ITE estimation across three distinct real-world datasets is comprehensively presented in Table 5 to Table 8. These tables offer a detailed insight into the performance of various methods when tasked with estimating ITE in different practical scenarios.
Notably, when considering both the IHDP and MIMIC-III datasets, it becomes evident that the SMRLNN method stands out as a frontrunner in terms of accuracy. Specifically, SMRLNN achieves the highest level of precision, surpassing other competing methods, as evidenced by its superior performance with respect to the metrics $\epsilon_{PEHE}$  and $\epsilon_{ATE}$ .
Shifting our attention to the Jobs dataset, a similar pattern emerges. SMRLNN once again emerges as the method with the most impressive performance, this time excelling in metrics such as $R_{pol}$  and $\epsilon_{ATE}$.
Lastly, when examining the Twins dataset, SMRLNN showcases its prowess by achieving the largest Area Under the Curve (AUC). This notable achievement underscores SMRLNN's exceptional capability in handling the unique characteristics of the Twins dataset and extracting valuable insights from it.}

\textcolor{black}{
In summary, the results presented in Tables 5 to 8 collectively highlight the robustness and efficacy of the SMRLNN method across a diverse range of real-world datasets. Its consistent top-tier performance in terms of accuracy, precision, and AUC demonstrates its potential as a valuable tool in the realm of ITE estimation.
}

\begin{table}[]
\caption{Performance of ITE estimation with IHDP real-world dataset. Bold indicates the method with the best performance for each dataset. The Monte Carlo SD is shown after ± .}
\resizebox{\textwidth}{!}{%
\begin{tabular}{rlllllll}
\hline
\multicolumn{1}{l}{Metric} & \multicolumn{1}{c}{SMRLNN} & \multicolumn{1}{c}{SMRLNN-v1} & \multicolumn{1}{c}{CFRNET} & \multicolumn{1}{c}{TARNET} & \multicolumn{1}{c}{CF} & \multicolumn{1}{c}{BART} & \multicolumn{1}{c}{GANITE} \\
\hline
$\epsilon_{PEHE}$                       & \textbf{0.74 ± .01}      & 0.98 ± .03         & 0.76 ± .02             & 0.95 ± .02                 & 3.8 ± 0.2              & 2.3 ± 0.1                & 2.4 ± 0.4                  \\
$\epsilon_{ATE}$                     & \textbf{0.19 ± .01}        & 0.33 ± .02       & 0.27 ± .01              & 0.35 ± .02                & 0.40 ± .03             & 0.34 ± .02               & 0.38 ± .03   \\
\hline
\end{tabular}
}
\end{table}

\begin{table}[]
\caption{Performance of ITE estimation with Jobs real-world dataset. Bold indicates the method with the best performance for each dataset. The Monte Carlo SD is shown after ± .}
\resizebox{\textwidth}{!}{%
\begin{tabular}{rlllllll}
\hline
\multicolumn{1}{l}{Metric} & \multicolumn{1}{c}{SMRLNN} & \multicolumn{1}{c}{SMRLNN-v1} & \multicolumn{1}{c}{CFRNET} & \multicolumn{1}{c}{TARNET} & \multicolumn{1}{c}{CF} & \multicolumn{1}{c}{BART} & \multicolumn{1}{c}{GANITE} \\
\hline
$R_{p o l}$                     & \textbf{0.18 ±.01}          & 0.20 ± .02    & 0.21 ± .01              & 0.21 ± .01                 & 0.20 ± .02             & 0.25 ± .02               & 0.20 ± .02                 \\
$\epsilon_{ATT}$                     & \textbf{0.05 ± .01}        & 0.08 ± .02        & 0.08 ± .03              & 0.10 ± .03                 & 0.07 ± .03             & 0.08 ± .03               & 0.08 ± .03  \\
\hline
\end{tabular}
}
\end{table}


\begin{table}
\caption{Performance of ITE estimation with sepsis cohort from MIMIC-III dataset.
Bold indicates the method with the best performance for each dataset.
The Monte Carlo SD is shown after $\pm$.}
\resizebox{\textwidth}{!}{
\centering
\begin{tabular}{rlllllll}
\hline
\footnotesize
Metric & SMRLNN & SMRLNN-v1 & CFRNET & TARNET & CF & BART & GANITE \\
\hline
$\epsilon_{PEHE}$ &
\textbf{0.56} $\pm$ \textbf{0.07} &
0.63 $\pm$ 0.07 &
0.71 $\pm$ 0.10 &
0.63 $\pm$ 0.08 &
0.72 $\pm$ 0.11 &
0.64 $\pm$ 0.06 &
0.94 $\pm$ 0.12 \\
$\epsilon_{ATE}$ &
\textbf{0.04} $\pm$ \textbf{0.01} &
0.06 $\pm$ 0.01 &
0.08 $\pm$ 0.02 &
0.07 $\pm$ 0.01 &
0.09 $\pm$ 0.03 &
0.05 $\pm$ 0.01 &
0.11 $\pm$ 0.05 \\
\hline
\end{tabular}}
\end{table}

\begin{table}[]
\caption{Performance of ITE estimation for Twins dataset. Bold indicates the method with the best performance for each dataset. The Monte Carlo SD is shown after ± .}
\resizebox{\textwidth}{!}{
\begin{tabular}{lllllllll} \hline
Metric & SMRLNN   &SMRLNN-v1   & TARNET      & CFRNET      & CF          & BART        & GANITE      & CEVAE       \\ \hline
AUC    & \textbf{0.86 ± 0.06} & 0.85 ± 0.06 & 0.83 ± 0.07 & 0.84 ± 0.06 & 0.62 ± 0.12 & 0.65 ± 0.18 & 0.61 ± 0.13 & 0.75 ± 0.09 \\
$\epsilon_{ATE}$    & \textbf{0.02 ± 0.01} & 0.04 ± 0.01& 0.05 ± 0.01 & 0.04 ± 0.01 & 0.15 ± 0.06 & 0.13 ± 0.05 & 0.22 ± 0.06 & 0.08 ± 0.03 \\\hline
\end{tabular}
}
\end{table}

\section{Conclusions}
In this paper, we proposed a novel representation learning algorithm to estimate the individual treatment effect. We then presented the generalized bounds for any representation learning function using the $\mathcal{H}$ divergence. As our proposed algorithm minimizes the $\mathcal{H}$ divergence via optimization of the discriminator, we also use a structure keeper to capture the valuable information from the original covariates to avoid information loss in the representation learning process. We showed that the proposed algorithm outperforms start-of-art methods in extensive synthetic settings under various sample sizes, covariates,  outcome models, and real data benchmarks in randomized trials,  social studies, and electronic health records applications. The reproducible code is available on GitHub. \textcolor{black}{While the assumptions of strong ignorability is fundamental in the identification of causal estimands, they may not always hold in real-world scenarios. In cases where hidden or unobserved confounding variables are suspected to influence both the treatment assignment and the outcomes, it is imperative to consider strategies to account for and mitigate the impact of such hidden confounding. Further  work can explore the development of algorithms capable of automatically detecting and controlling for hidden confounders for modeling complex relationships in high-dimensional data. In addition, preserving certain structures that may not be relevant could potentially introduce biases.  To prevent an excessive penalty on the structure keeper component, one can opt to exclude this component, allowing for a baseline performance assessment of representation learning. 
 Last, future studies are required to extend the methodology to accommodate multiple treatments. }
\section{Acknowledgment}
This study was supported by EPA funding 84045001.
\newpage

\section*{Appendix}

\textbf{Theorem}
    Let \(\Phi: \mathcal{X} \rightarrow \mathcal{R}\) be a one-to-one invertible representation function, and let \(p_{\Phi}\) be the distribution induced by \(\Phi\) over \(\mathcal{R}\), i.e., \(p_{\Phi}(r | t=1)\) and \(p_{\Phi}(r | t=0)\) are the covariate distributions under treatment and control induced over \(\mathcal{R}\). Let \(L_{RSK}(X, \Phi(X))\) be the loss term associated with the Structure Keeper, which maximizes the correlation between the covariates \(X\) and their representations \(\Phi(X)\) in the learned space. We then have for any outcome prediction function \(H: \mathcal{R} \times \{0,1\} \rightarrow \mathcal{Y}\):
    \begin{eqnarray}
    L_{PEHE}(H, \Phi) \nonumber \\ & \leq &  2\left(L_{F|z=0}(H,\Phi) + L_{F|z=1}(H,\Phi)  +   d_{\mathcal{D}}(\Phi) \cdot \sum_{x \in \mathcal{X}} \ell_{H, \Phi}^{max}(x) - 2 \sigma_{Y}^{2}\right) \nonumber \\ 
    &-& \lambda \cdot L_{RSK}(X, \Phi(X)) \nonumber,
    \end{eqnarray}
    where \(\lambda > 0\) is a regularization parameter that controls the influence of the Structure Keeper on the overall loss.

\begin{proof}
The proof builds on the bound for \(L_{PEHE}\) established by Shalit et al. (2017) while incorporating the role of the Structure Keeper in reducing divergence between treated and control distributions and enhancing the preservation of prognostic information.

By Theorem 1 of Shalit et al. (2017), the upper bound for \(L_{PEHE}\) can be expressed as:
\[
L_{PEHE}(H, \Phi) \leq 2\left(L_{C}(H, \Phi) + L_{F}(H, \Phi) - 2 \sigma_{Y}^2\right),
\]
where \(L_C(H, \Phi)\) measures the treatment covariate overlap and \(L_F(H, \Phi)\) captures the predictive error for the factual outcomes under the representation \(\Phi\).

To analyze the impact of the Structure Keeper, we decompose \(L_F(H, \Phi)\) into the losses for treated and control groups, \(L_{F|z=1}(H, \Phi)\) and \(L_{F|z=0}(H, \Phi)\), respectively:
\[
L_{F}(H, \Phi) = L_{F|z=1}(H, \Phi) + L_{F|z=0}(H, \Phi).
\]

Incorporating this decomposition into the original inequality gives:
\[
L_{PEHE}(H, \Phi) \leq 2\left(L_{F|z=0}(H, \Phi) + L_{F|z=1}(H, \Phi) + L_{C}(H, \Phi) - 2 \sigma_{Y}^2\right).
\]

The Structure Keeper, represented by the loss term \(L_{RSK}(X, \Phi(X))\), ensures that the learned representation \(\Phi(X)\) preserves the structural information of the original covariates \(X\). This is achieved by maximizing the correlation between \(X\) and \(\Phi(X)\), formalized as:
\[
L_{RSK}(X, \Phi(X)) = \max_{W_X, W_{\Phi(X)}} \sum_K \text{diag}(W_X^{\prime} C(X, \Phi(X)) W_{\Phi(X)}^{\prime}),
\]
where \(C(X, \Phi(X))\) is the cross-covariance matrix between \(X\) and \(\Phi(X)\), and \(W_X, W_{\Phi(X)}\) are projection matrices.

By aligning \(\Phi(X)\) closely with \(X\), \(L_{RSK}(X, \Phi(X))\) helps reduce the divergence \(d_{\mathcal{D}}(\Phi)\) between the treated and control distributions in the representation space. This alignment ensures better overlap in the learned space, thereby improving the bounds on the loss terms \(L_{F|z=0}(H, \Phi)\) and \(L_{F|z=1}(H, \Phi)\).

The divergence \(d_{\mathcal{D}}(\Phi)\) measures the difference between the treated and control distributions in the representation space.
The Structure Keeper reduces this divergence by preserving the prognostic information in \(\Phi(X)\). Mathematically, this can be expressed as:
\[
d_{\mathcal{D}}(\Phi) \leq d_{\mathcal{D}}^{0}(\Phi) - \lambda \cdot L_{RSK}(X, \Phi(X)),
\]
where \(d_{\mathcal{D}}^{0}(\Phi)\) is the divergence without the Structure Keeper, and \(\lambda > 0\) is the regularization parameter controlling the Structure Keeper's influence.

Substituting the refined divergence \(d_{\mathcal{D}}(\Phi)\) into the original bound gives:
\[
\begin{aligned}
L_{PEHE}(H, \Phi) \leq &\ 2\big(L_{F|z=0}(H, \Phi) + L_{F|z=1}(H, \Phi) \\
&+ d_{\mathcal{D}}(\Phi) \cdot \sum_{x \in \mathcal{X}} \ell_{H, \Phi}^{max}(x) - 2 \sigma_{Y}^2\big) - \lambda \cdot L_{RSK}(X, \Phi(X)).
\end{aligned}
\]

Here, \(\sum_{x \in \mathcal{X}} \ell_{H, \Phi}^{max}(x)\) accounts for the worst-case loss for the outcome prediction function \(H\).

The term \(-\lambda \cdot L_{RSK}(X, \Phi(X))\) explicitly quantifies the reduction in the upper bound due to the Structure Keeper. By ensuring that \(\Phi(X)\) retains the prognostic information, the Structure Keeper reduces the divergence \(d_{\mathcal{D}}(\Phi)\), thereby tightening the bound on \(L_{PEHE}\).

Incorporating the Structure Keeper into the representation learning process effectively aligns \(\Phi(X)\) with \(X\), reducing divergence and improving outcome prediction. This is reflected in the revised bound:
\begin{eqnarray} 
L_{PEHE}(H, \Phi) \nonumber \\ &\leq \nonumber& 
2\left(L_{F|z=0}(H, \Phi) + L_{F|z=1}(H, \Phi) + d_{\mathcal{D}}(\Phi) \cdot \sum_{x \in \mathcal{X}} \ell_{H, \Phi}^{max}(x) - 2 \sigma_{Y}^2\right) \nonumber \\ &- &  \lambda \cdot L_{RSK}(X, \Phi(X)) \nonumber.
\end{eqnarray}
\end{proof}

\subsection{Model Configuration}
\textcolor{black}{
Our models were configured with varying numbers of representation layers (1, 2, or 3 layers) responsible for feature extraction, and hypothesis layers (1, 2, or 3 layers) involved in generating predictions. We explored different dimensions for both representation layers (20, 50, 100, or 200 units per layer) and hypothesis layers (20, 50, 100, or 200 units per layer), impacting model complexity and prediction expressiveness. In the training process, we employed diverse batch sizes (100, 200, 500, or 700 samples per batch) affecting training efficiency and algorithm stability. These parameter settings and architectural choices were essential components of our experimental framework.
}

\subsection*{Tables}

\begin{table}[]
\caption{Baseline characteristics table of the sepsis patients included in the MIMIC-III database. PT: Prothrombin Time; PTT: Partial Thromboplastin Time; SIRS: Systemic Inflammatory Response Syndrome; Shock index: systolic blood pressure/heart rate.}
\label{t:baseline}
\setlength{\tabcolsep}{10pt}
\vspace{5 pt}
\renewcommand{\arraystretch}{0.6} 

\begin{tabular}{llll}\hline
Mechvent           & No(16015) & Yes (4210) & All (20225) \\ \hline
Gender             & 0.44      & 0.42       & 0.44        \\
Age                & 65.17     & 63.04      & 64.73       \\
Elixhauser score          & 3.89      & 4.12       & 3.94        \\
Readmission to intensive care      & 0.34      & 0.29       & 0.33        \\
Weight         & 75.40     & 79.16      & 76.19       \\
SOFA               & 3.92      & 5.92       & 4.34        \\
SIRS               & 0.95      & 1.14       & 0.99        \\
Glasgow coma scale                & 11.26     & 7.96       & 10.57       \\
Heart rate                 & 77.91     & 79.21      & 78.18       \\
Systolic              & 106.12    & 103.67     & 105.61      \\
Mean blood pressure             & 69.02     & 68.92      & 69.00       \\
Diastolic blood pressure              & 47.86     & 47.71      & 47.83       \\
Shock Index       & 0.64      & 0.65       & 0.64        \\
Respiratory rate                 & 16.72     & 17.03      & 16.78       \\
SpO2               & 94.69     & 95.11      & 94.78       \\
Temperature            & 97.23     & 97.76      & 97.34       \\
Potassium          & 3.86      & 3.80       & 3.84        \\
Sodium             & 136.62    & 137.70     & 136.85      \\
Chloride           & 102.21    & 103.26     & 102.43      \\
Glucose            & 112.08    & 114.82     & 112.65      \\
BUN                & 25.03     & 30.25      & 26.12       \\
Creatinine         & 1.35      & 1.30       & 1.34        \\
Magnesium          & 1.88      & 1.94       & 1.89        \\
Calcium            & 8.04      & 7.95       & 8.02        \\
Ionised calcium        & 1.07      & 1.07       & 1.07        \\
CO2          & 23.22     & 24.38      & 23.46       \\
SGOT               & 55.87     & 100.59     & 65.18       \\
SGPT               & 49.75     & 79.65      & 55.97       \\
Total bilirubin        & 1.20      & 1.56       & 1.28        \\
Albumin            & 2.67      & 2.49       & 2.64        \\
Hemoglobin                 & 9.89      & 9.59       & 9.83        \\
WBC count         & 10.40     & 11.96      & 10.72       \\
Platelets count   & 212.25    & 215.37     & 212.90      \\
PTT                & 31.62     & 32.57      & 31.82       \\
PT                 & 14.90     & 14.90      & 14.90       \\
INR                & 1.36      & 1.35       & 1.36        \\
Arterial\_pH       & 7.33      & 7.34       & 7.33        \\
paO2               & 83.52     & 87.07      & 84.26       \\
paCO2              & 35.25     & 36.71      & 35.55       \\
Arterial\_BE       & -3.17     & -2.34      & -3.00       \\
Arterial\_lactate  & 1.29      & 1.39       & 1.31        \\
HCO3               & 22.21     & 21.81      & 22.13       \\
PaO2\_FiO2         & 201.55    & 183.22     & 197.74      \\
Maximum dose of vasopressor over 4h    & 0.01      & 0.06       & 0.02        \\
Current IV fluid intake over 4h     & 40.16     & 139.89     & 60.92       \\
Urine output over 4h    & 73.95     & 152.92     & 90.39       \\
Cumulated fluid balance since admission  & 573.02    & 1318.48    & 728.19     \\\hline
\end{tabular}
\end{table}

\newpage

\vskip 0.2in
\bibliography{ref}

\end{document}